\documentclass[letterpaper, 10pt, conference]{ieeeconf}  % Comment this line out if you need a4paper

\usepackage{times}  % DO NOT CHANGE THIS
\usepackage{helvet}  % DO NOT CHANGE THIS
\usepackage{courier}  % DO NOT CHANGE THIS
\usepackage[hyphens]{url}  % DO NOT CHANGE THIS
\usepackage{graphicx} % DO NOT CHANGE THIS
\usepackage{caption} % DO NOT CHANGE THIS AND DO NOT ADD ANY OPTIONS TO IT

\usepackage{microtype}
\usepackage{graphics}
\usepackage{subcaption}
\usepackage{booktabs} % for professional tables
\usepackage{xcolor}

\usepackage{hyperref}
\usepackage{cite}

\usepackage{makecell}
\usepackage{fontawesome}

\usepackage{epstopdf}
\usepackage{rotating,multirow}
\usepackage{amsmath,amssymb} % define this before the line numbering.
\usepackage{color}
\usepackage{wrapfig,lipsum}
\usepackage{tabularx}
\usepackage{booktabs, xcolor, subcaption, multirow, tikz}
\usepackage{algorithm}
\usepackage{algorithmic}
\usepackage[algo2e]{algorithm2e}
\usepackage{mathtools}

\newcommand{\etal}{\textit{et al. }}

\newcommand{\ie}{\textit{i.e., }}

\makeatletter
\newcommand*\bigcdot{\mathpalette\bigcdot@{.5}}
\newcommand*\bigcdot@[2]{\mathbin{\vcenter{\hbox{\scalebox{#2}{$\m@th#1\bullet$}}}}}
\makeatother
\newenvironment{smalltable}[1]{ \footnotesize \begin{tabular}{#1} }{ \end{tabular} }

\usepackage{newfloat}
\usepackage{listings}

\usepackage{xcolor}

\IEEEoverridecommandlockouts                              % This command is only needed if 
\title{\LARGE \bf Multi-class Road Defect Detection and Segmentation using \\ Spatial and Channel-wise Attention for Autonomous Road Repairing}

\author{Jongmin Yu$^{1,2}$, Chen Bene Chi$^{3}$, Sebastiano Fichera$^{4,5}$,\\ Paolo Paoletti$^{4,5}$, Devansh Mehta$^{4}$, and Shan Luo$^{2,\dagger}$% <-this % stops a space
\thanks{$^{1}$Department of Applied Mathematics and Theoretical Physics, University of Cambridge, Wilberforce Rd, Cambridge CB3 0WA, United Kingdom {\tt\small jy522@cam.ac.uk} (was affiliated with Department of Engineering, King's College London, Strand, London, WC2R 2LS, United Kingdom).}
\thanks{$^{2}$Department of Engineering, King's College London, Strand, London, WC2R 2LS, United Kingdom, {\tt\small shan.luo@kcl.ac.uk}.}
\thanks{$^{3}$Department of Informatics, King's College London, Strand, London, WC2R 2LS, United Kingdom, {\tt\small chi.b.chen@kcl.ac.uk}.}
\thanks{$^{4}$Robotiz3D, Sci-Tech Daresbury Keckwick Lane, WA4 4FS, United Kingdom, {\tt\small  \{devansh.mehta\}@robotiz3d.com}}
\thanks{$^{5}$School of Engineering, University of Liverpool, Brownlow Hill, L69 3GH, Liverpool, United Kingdom, {\tt\small \{sebastiano.fichera, paoletti\}@liverpool.ac.uk}}
\thanks{This work was supported by the Innovate UK SMART grant ``ARRES PREVENT: The World-First Autonomous Road Repair Vehicle'' (10006122).}
\thanks{$\dagger$ represents the corresponding author.}
}
\begin{document}

\maketitle
\thispagestyle{empty}
\pagestyle{empty}

\begin{abstract}
Road pavement detection and segmentation are critical for developing autonomous road repair systems. However, developing an instance segmentation method that simultaneously performs multi-class defect detection and segmentation is challenging due to the textural simplicity of road pavement image, the diversity of defect geometries, and the morphological ambiguity between classes. We propose a novel end-to-end method for multi-class road defect detection and segmentation. The proposed method comprises multiple spatial and channel-wise attention blocks available to learn global representations across spatial and channel-wise dimensions. Through these attention blocks, more globally generalised representations of morphological information (spatial characteristics) of road defects and colour and depth information of images can be learned. To demonstrate the effectiveness of our framework, we conducted various ablation studies and comparisons with prior methods on a newly collected dataset annotated with nine road defect classes. The experiments show that our proposed method outperforms existing state-of-the-art methods for multi-class road defect detection and segmentation methods.
\end{abstract}
%Doing reivison
\section{Introduction}
\label{sec:1}
Bad road conditions pose multiple problems: they contribute to vehicle damage, cause nearly 13\% of car accidents in the UK \cite{uk2022caraccident}, increase carbon emissions \cite{Wang2020roadcarbon}, and lead to higher fuel consumption \cite{Alsaadi2021roadcarbon}. Currently, human workers usually handle road repairs, which is both dangerous and inefficient. Emerging technologies are looking to employ machines powered by advanced computer vision algorithms. These algorithms use object detection and segmentation techniques to accurately identify and mark road defects. Proper road defect detection ensures that the repair machines can locate precisely where the issues are, while segmentation helps in understanding the extent of the defect, thereby optimising the amount of repair material needed \cite{katsamenis2022robotic,eskandariapplication}. Therefore, developing precise object detection and segmentation methods is crucial for making these automated road repair systems effective and cost-efficient.

% Road surface imperfections play a notable role in causing vehicle damage and occasionally lead to accidents, compromising the safety of travellers. Data from the UK's Transport Department \cite{uk2022caraccident} reveals that about 12.6\% of car mishaps in 2020 were due to inadequate road conditions. Moreover, recent studies \cite{Wang2020roadcarbon,Alsaadi2021roadcarbon} suggest that poor road conditions can boost carbon emissions by roughly 2\% and raise fuel consumption by about 5\%. Accordingly, repairing road defects is an important task; However, now a day, human workers usually operate road defect repair, and their working environment is hazardous and poor. Various automatic road repair systems have been proposed to alternate the human workers \cite{katsamenis2022robotic,eskandariapplication}.

% In developing autonomous road repairing systems, developing precise road defect detection and segmentation methods is a critical mission because these methods significantly impact how the system locates the defects on the road and how much repair material would be used. Errors in the detection and segmentation methods will cause inefficiency in the system by wasting the operation time and the repair materials. As a result, developing accurate road defect detection and segmentation methods is an essential challenge in improving the performance and cost efficiencies of the system. 

\begin{figure}[!t]
	\centering
	\includegraphics[width=\columnwidth,height=5cm]{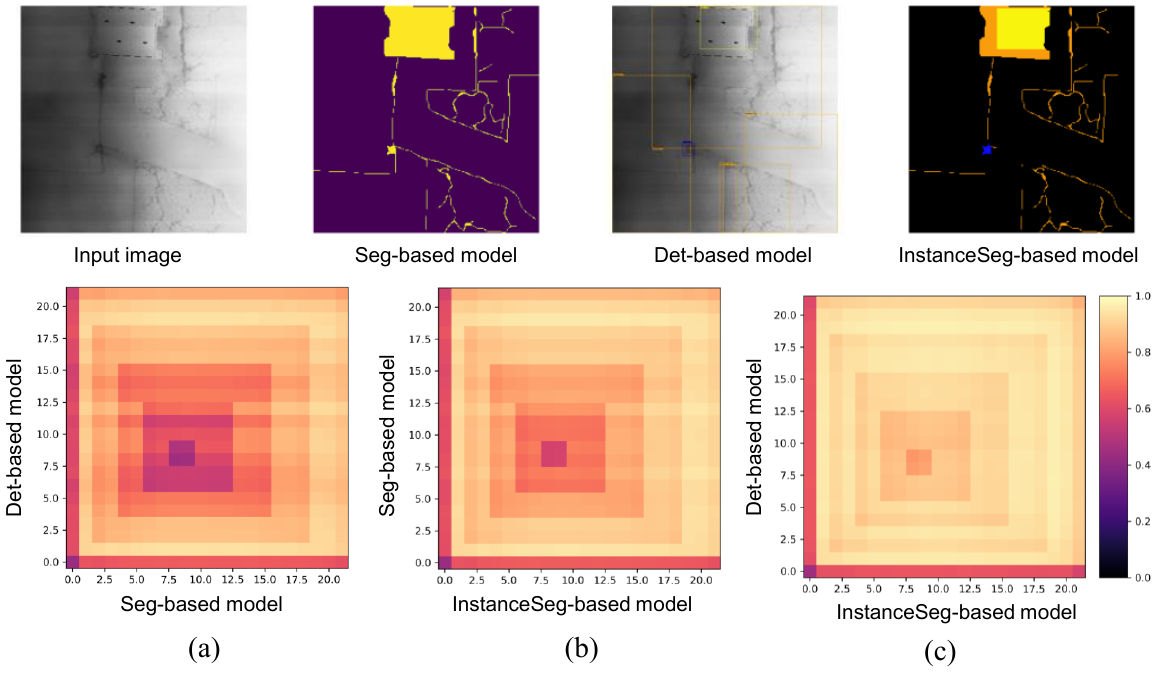}
	\caption{Visualisation of Kernel similarities computed based on Centered Kernel Alignment (CKA) \cite{kornblith2019similarity}. (a) denotes the kernel similarity map for segmentation (Seg-based model) and detection (Det-based model) models. (b) denotes the kernel similarity map for segmentation and instance segmentation (InstanceSeg-based model) models. (c) denotes the kernel similarity map for detection and instance segmentation models. Each axis denotes the depth of layers. The brighter the colour, the higher the kernel similarity.}
	\label{fig:1}
	\vspace{-2ex}
\end{figure}

Numerous methods have been proposed for road defect segmentation, showing considerable success for publicly available datasets \cite{li2018convolutional,yu2018joint,yu2020unsupervised,fan2019road}. It can be thought that methodological advancements alongside deep learning technologies led to these achievements. However, it can not be denied that the distinct colours and textures of road defects are compared to normal surfaces.
%However, it can not be denied that the distinct colours and textures of road defects are compared to standard surfaces so that this distinction makes the easy to classify road defects. 
Multi-class road defect detection still faces challenges because of diverse defect geometries and morphological ambiguities between classes. While recent deep learning methods have shown promise \cite{arya2021rdd2020,zhang2020exploring,wan2022yolo}, they struggle with limitations such as detecting defects at the image unit level (\ie distinguishing whether defects exist in an image) \cite{zhang2020exploring} or focusing on defect classes that are easy to be identified  \cite{arya2021rdd2020,wan2022yolo}. To perform the segmentation and detection simultaneously, some studies utilise instance segmentation approaches \cite{he2022pavement,zhang2022multi,dong2019pga}. However, unlike general approaches of instance segmentation  \cite{he2017mask,liang2022cbnet,lee2020centermask}, it takes the form of a two-stage framework that crops images after detection and then proceeds with segmentation for each image.

In addressing road defect segmentation and detection within a single framework, the network must adeptly learn colour attributes useful for segmentation and spatial information crucial for multi-class detection. Based on the Mask-RCNN \cite{he2017mask} structure, we analyse the kernel similarities measured by Centered Kernel Alignment (CKA) \cite{kornblith2019similarity} between three models trained by segmentation, detection, and instance segmentation objectives. We find out that the model trained by an instant segmentation objective is more similar to the model trained by a detection objective (See Fig. \ref{fig:1}, the colour of kernel similarity map between the detection model and the instance segmentation model (Fig. \ref{fig:1}(c)) is lighter than the kernel similarity map between the instant model and segmentation model (Fig. \ref{fig:1}(b))). This indicates that the learned kernels may be biased in the detection information. As a result, these disparities have necessitated a two-stage framework design in prior studies \cite{he2022pavement,zhang2022multi,dong2019pga}. However, this two-stage approach undermines computational efficiency since the computational time for the segmentation depends on detection results.

To overcome this, we introduce a novel Spatial and Channel-wise Multi-head Attention Mask-RCNN (SCM-MRCNN) for detecting and segmenting multi-road defect classes. Built upon the foundation of MRCNN, SCM-MRCNN incorporates Spatial and Channel-wise Multi-head Attention (SCM-attention) blocks. These blocks are designed to extract latent features broadly robust to variations in geometric shape and colour information of road defects within a given image. Not only is the attention mechanism applied across the spatial domain, but similar techniques are applied to the channel axis to extract features with enhanced generalisation of both the spatial and channel axes. Ablation studies show that SCM attention improves multi-class defect detection and segmentation performance.

Additionally, we introduce a new dataset named the RoadEYE to provide a new benchmark for multi-class road defect detection and segmentation. Unlike existing road defect detection datasets \cite{arya2021rdd2020} that only provide detection labels in the form of bounding boxes for multiple classes, or other general segmentation datasets \cite{zou2012cracktree} that consider all defects as a single class without class distinction, the proposed dataset provides bounding for multi-class road defects. We provide a new benchmark for multi-class defect detection and segmentation research by providing box-shaped and segmentation labels for six classes of road defects. We conduct extensive experiments to demonstrate the proposed SCM-Attention block's efficacy and evaluate its performance in multi-class road defect detection and segmentation. The SCM-MRCNN achieves a mean average precision (mAP) of 61.7\% on the RDD2020 dataset \cite{arya2021rdd2020}. For defect segmentation, the SCM-MRCNN achieves an average intersection over union (AIU) of 53.1\% on the Surface Crack dataset \cite{zou2012cracktree}. Experimental results on the RoadEYE dataset indicate that SCM-MRCNN outperforms the state-of-the-art (SOTA) method \cite{bolya2019yolact} by 1.8\% in multi-class road defect detection and segmentation.

Our key contributions are as follows:
\begin{itemize}
    \item A spatial and channel-wise multi-head attention Mask-RCNN (SCM-MRCNN), the first unified framework for multi-class road defect detection and segmentation.
    \item The proposed Spatial and Channel-wise attention block can learn more global representation for both spatial and channel-wise dimensions.
    \item A new multi-class road defect detection and segmentation dataset (RoadEYE dataset) is collected and annotated. 
\end{itemize}

%The rest of this paper is organised as follows. We present related works in Section \ref{sec:2}. Section \ref{sec:3} introduces the proposed framework and the SC-Attention block. We describe our experimental settings and results in Section \ref{sec:4}, with ablation studies and comparisons with existing SOTA methods. We conclude this paper in Section \ref{sec:5}. 

\section{Related works} 
\label{sec:2}

Before 2008, defect detection relied on signal filtering~\cite{li_novel_2008,tanaka_crack_1998} using pixel values, assuming ``cracks are deeper than surroundings". The algorithm differentiated cracks and roads via these methods, achieving 95.5\% success in feature-based filtering\cite{tanaka_crack_1998,maode_pavement_2007,sy_detection_2008}. This involved pre-defined defect descriptions and layered thresholding. Yet, these methods have limitations. They're adept mainly at detecting one specific type of crack, often struggling with noise and differentiating cracks from similar-shaped objects, especially in low-contrast settings. Adapting them for new tasks or defects is challenging due to their reliance on thresholding-defined crack features \cite{nguyen_automatic_2009}, necessitating significant rework.

Machine learning techniques, when combined with handcrafted features, address road image classification challenges. These techniques utilise various extractors to discern road defects. \cite{a_supervised_2009} They are more adaptable than traditional methods, with the suitable feature extractor producing solid results. \cite{cord_automatic_2012} Using varied feature extractors and Adaboost enhances defect detection on multiple surfaces. While traditional algorithms like Random Forest are utilised\cite{breiman_random_2001}, SVM has become prominent recently due to its noise-reduction capabilities. \cite{prasanna_automated_2016} However, these methods face challenges. The effectiveness of hand-crafted features can significantly influence performance. Moreover, traditional machine learning cannot often extract high-level features efficiently.

% The discovery of deep learning revolutionised feature extraction compared to traditional machine learning. Deep learning automatically extracts features in a stable and generalised manner, reducing the effort needed to adapt existing systems for new tasks. Early deep learning models like Artificial Neural Networks (ANNs) detected image cracks but were limited by computational demands \cite{kaseko_neural_1993,liu_detection_2002}. Convolution Neural Networks (CNNs) and variations emerged, enabling efficient whole-image processing in computer vision. Multi-layer CNN like FCN, incorporating multi-scale feature extraction, further improved performance \cite{zhong_cnn-based_2019}.
% CNNs instigated a paradigm shift but have inherent limitations. They often cannot discern local pixel nuances and global context simultaneously. Moreover, they grapple with class imbalances, leading to biases towards over-represented classes, making them less fit for multi-class detection and segmentation. Thus, researchers proposed other solutions that come from CNN.

Deep learning revolutionised feature extraction compared to traditional machine learning, automatically extracting features in a stable and generalised manner, reducing adaptation effort for new tasks. Early models like Artificial Neural Networks (ANNs) detected image cracks but were computationally demanding\cite{kaseko_neural_1993,liu_detection_2002}. Convolution Neural Networks (CNNs) enable efficient whole-image processing in computer vision. Multi-layer CNNs like FCN, with multi-scale feature extraction\cite{zhong_cnn-based_2019} or multiple labelled source domains\cite{yu2023multi}, improved performance. However, CNNs have limitations: they simultaneously struggle with pixel nuances and global context and face class imbalance issues, making them less suitable for multi-class tasks. Therefore, researchers proposed deep-learning based solutions.

To address road defect segmentation, methods combine CNN and attention mechanism proposed\cite{qiao_automatic_2021, ong_feature_2023, konig_convolutional_2019}. Attention mechanisms enable the network to emphasise crucial parts and reduce the weight of background pixels. Transformers, using self-attention mechanisms\cite{vaswani_attention_2023}, are employed for segmentation. Several methods integrate CNN with the transformer, achieving high performance\cite{liu_crackformer_2021,guo_pavement_2023, asadi_shamsabadi_vision_2022}. However, concerns have been raised about their inference speed \cite{wang_automatic_2022}, necessitating extra effort to utilise these solutions for instant road defect segmentation.
% On the other hand, RCNN (Region-Based Convolutional Neural Network) \cite{6909475} addresses object detection and classification. Detecting object regions is challenging. The RCNN utilised 2,000 region proposals, slowing down real-time results. 
% Later versions like Fast-RCNN\cite{girshick_fast_2015} and Faster-RCNN\cite{ren_faster_2017,xu_crack_2022,wang_road_2018,li_automatic_2021} reduced training time and moved towards real-time road defect detection by integrating automatic region-proposing methods. Mask-RCNN\cite{he2017mask} adds object segmentation capabilities on top of faster RCNN.
% % Mask-RCNN\cite{he2017mask} adds object detection, classification, and image masking capabilities.

In the domain of road defect detection and classification, the introduction of RCNN (Region-Based Convolutional Neural Network) \cite{6909475} represented a notable breakthrough, albeit with the drawback of relying on an extensive 2,000 region proposal process that impeded real-time performance. Subsequent iterations like Fast-RCNN \cite{girshick_fast_2015} and Faster-RCNN \cite{ren_faster_2017,xu_crack_2022,wang_road_2018,li_automatic_2021} addressed this limitation by integrating automatic region-proposing methods, significantly reducing training time, and advancing towards real-time road defect detection. Mask-RCNN \cite{he2017mask} further extends Faster-RCNN by introducing object segmentation capabilities, enhancing its suitability for instant road defect detection and segmentation.

\begin{figure*}[ht] 
  \begin{subfigure}{0.49\linewidth}
    \includegraphics[width=\linewidth]{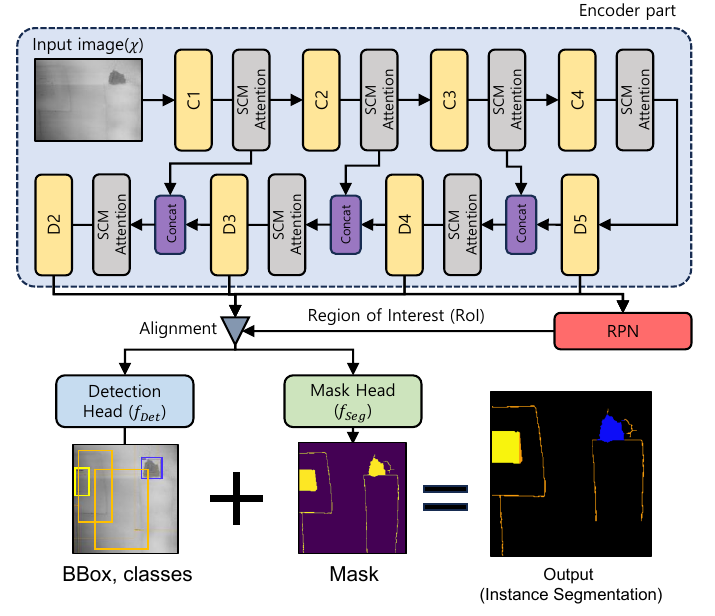}
    \caption{}\label{subfig:key-a}
  \end{subfigure}\hfill
  \begin{minipage}{0.5\linewidth}
  \vspace{-50ex}
    \begin{subfigure}{\linewidth}
      \includegraphics[width=\linewidth]{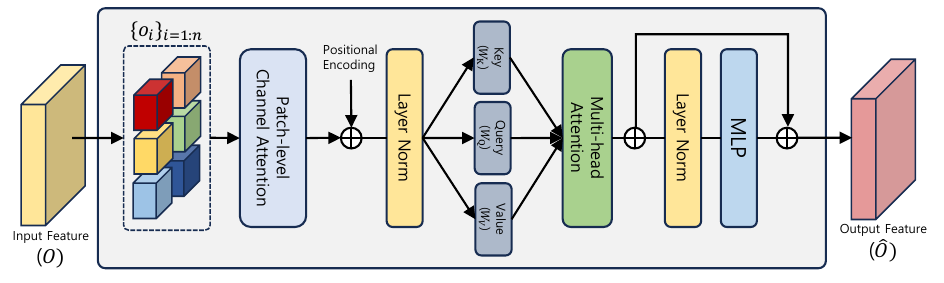}
      \caption{}\label{subfig:key-b}
    \end{subfigure}\hfill
    \vspace{4ex}
    \begin{subfigure}{\linewidth}
      \includegraphics[width=\linewidth]{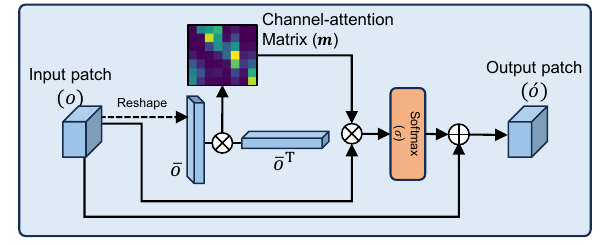}
      \caption{}\label{subfig:key-c}
    \end{subfigure}\hfill
  \end{minipage}
  \caption{Structural details of the proposed SCM-MRCNN and SCM attention block. (a) denotes Architectural details of the proposed SCM. C\# and D\# denote convolutional and deconvolutional layers, respectively. 'Concat' represents a concatenating operation between two latent features. (b) and (c) represents the structural details of the proposed spatial and channel-wise multi-head attention (SCM-attention) block and patch-based channel attention, respectively. $\otimes$ and $\oplus$ denote element-wise multiplication and element-wise addition, respectively.}
  \label{fig:example-key}
  \vspace{-2ex}
\end{figure*}

%\begin{figure}[!t]
%	\centering
%	\includegraphics[width=\columnwidth]{./figures/figure2.pdf}
%	\caption{Architectural details of the proposed SCM. C\# and D\# denote covolutional and deconvolutional layers, respectively. 'Concat' represents a concatenating operation between two latent features. The architectural details of the proposed SCM-RCNN are based on the FPN-based Mask RCNN \cite{he2017mask}.}
%	\label{fig:2}
%	\vspace{-0.2cm}
% \end{figure}

\section{The proposed method}
\label{sec:3}
\subsection{Method Overview}
As mentioned in Section \ref{sec:1} and shown in Fig. \ref{fig:1}, General MRCNN tend to be biased into detection objectives regarding learning a  multi-class road defect detection and segmentation method. The model should learn spatially and channel-wise robust representation to produce a good segmentation and detection performance using a single framework. As illustrated in Fig. \ref{subfig:key-a}, the architecture of our proposed SCM-MRCNN is inspired by MRCNN \cite{he2017mask}. SCM-attention block is placed between the convolutional and deconvolutional layers. The convolutional and deconvolutional layers are used to down-sample or up-sample the features. SCM-attention blocks have been located in between each convolutional or deconvolutional layer to reinforce spatial and channel-wise robustness of learnt features.

To learn spatially and channel-wisely more robust representation for multi-class road defect detection and segmentation, we propose a new spatial and channel-wise multi-head attention (SCM attention) block to fuse the advantage of the long dependency modelling in Transformer and channel-wise attention. The SCM attention block comprises two steps: patch-level channel attention and multi-head attention.  

\textbf{Patch-level Channel Attention:} Given the features extracted from each convolutional or deconvolutional layers $\mathbf{O} \in \mathbb{R}^{H\times{}W\times{}C_i},(i=1,2,3,..,n)$, we first make a set of small-size patch by splitting the features using pre-defined patch sizes $P$. This process is only conducted for spatial dimension so that we keep the original channel dimensions through this process $\mathbf{O} =\{o_{i}\}_{i=1:n_{p}}$, where $n_{p}$ is the number of patches. Then, we reshape each patch into a 2D flattened shape $\bar{o}_{i}\in\mathcal{R}^{hw\times{}C}$. We compute the channel-attention matrix $\mathbf{m}$ by multiplying the reshaped patch and fusing the matrix with the original patch $\mathbf{m}=\bar{o}_{i}\bar{o}^{\mathbf{T}}_{i}$. Channel-wise attention feature is finally obtained by applying the softmax function across the channel axes of the fused patch, and residual structure is applied to encode channel and dependencies for refining features from each latent feature, as follows: 
\begin{equation}
    \begin{split}
        \hat{o}_i  & = o_{i}+\sigma\left(\mathbf{m}o_{i}\right) \\ &=  o_{i}+\sigma\left(\bar{o}_{i}\bar{o}^{\mathbf{T}}_{i}o_{i}\right) 
    \end{split}
\end{equation}

%After that, we conduct residual operations with the attention feature with the original feature to prevent a shift-invariant issue \cite{he2016deep} and refine features \cite{liang2022cbnet}. 
Fig.~\ref{subfig:key-c} represents the detailed process of the patch-level channel attention. After finishing the patch-level channel attention, outputs are applied to the multi-head attention to learn spatially reinforced features. 

%fre

%\begin{figure}[!t]
%	\centering
%	\includegraphics[width=\columnwidth]{./figures/figure3.pdf}
%	\caption{Architectural details of the patch-based channel attention. $\otimes$ and $\oplus$ denote element-wise multiplication and element-wise addition, respectively.}
%	\label{fig:3}
%	\vspace{-0.2cm}
%\end{figure}

\textbf{Multi-head Attention:} We conduct the tokenisation of the outputs of the patch-level channel attention by reshaping the features into sequences of flattened 2D patches. Positional encoding is applied to each token. The tokens are then fed into the multi-head cross-attention layer (See Fig. \ref{subfig:key-b}), followed by a Multi-Layer Perceptron (MLP) with residual structure to encode channel and dependencies for refining features from each latent feature. 

Each token $\mathbf{T}_i$ is applied to multi-head attention by defining queries $\mathbf{Q}_i$, key $\mathbf{K}_i$, and value $\mathbf{V}_i$:
\begin{equation}
    \mathbf{Q}_i= \mathbf{T}_i W_{\mathbf{Q}_{i}},\mathbf{K}_{i}=\mathbf{T}_{i}W_{\mathbf{K}},\mathbf{V}_{i}=\mathbf{T}_{i}W_{\mathbf{V}}
\end{equation}
where $W_{\mathbf{Q}_{i}} \in \mathbb{R}^{C_i \times d}, W_\mathbf{K} \in \mathbb{R}^{C_i \times d}, W_\mathbf{V} \in \mathbb{R}^{C_i \times d}$ are kernels for mapping the key, value, and queries. $d$ denotes the sequence length (patch numbers).

With $\mathbf{Q}_{i} \in \mathbb{R}^{C_i \times d}, \mathbf{K} \in \mathbb{R}^{C_\Sigma \times d}, \mathbf{V} \in \mathbb{R}^{C_{\Sigma} \times d}$, the attention matrix $\mathbf{M}_i$ are produced and the value $\mathbf{V}$ is weighted by $\mathbf{M}_i$ through a self-attention mechanism:
\begin{equation}
    \begin{aligned}
        \mathrm{A}_i  = \mathbf{M}_i \mathbf{V^\top}  &= 
        \sigma\left[\psi\left(\frac {\mathbf{Q}^\top_i\mathbf{K}_{\Sigma}} {\sqrt{C_i}}\right)\right] \mathbf{V}^\top_{i} \\ 
    \end{aligned}
\end{equation}
where $\mathbf{K}_{\Sigma}$ defines a matrix storing the keys: $\mathbf{K}_{\Sigma}=\{K_{i}\}_{i=1:d}$ and $\mathbf{K}_{\Sigma}\in\mathcal{R}^{d\times{}d}$. $\psi(\cdot)$ and $\sigma(\cdot)$  denote the instance normalisation\cite{nam2018batch} and the softmax function, respectively.

In a multi-head attention with $N$-heads, the output is calculated as follows:
\begin{equation}
    \mathrm{MA}_i =(\mathrm{A}_{i}^1+ \mathrm{A}_{i}^2+,\dots,+\mathrm{A}_{i}^N) / N
\end{equation}

Hereinafter, applying an MLP and residual operator, the output is obtained as follows: 
\begin{equation}
    \mathbf{O}_i = \mathrm{{MA}}_i+ \mathrm{MLP}(\mathrm{MA}_i)
\end{equation}
For simplicity, we omitted the layer normalisation (See Layer Norm blocks in Fig \ref{subfig:key-b}) in equation descriptions. The operation of the SCM attention blocks can be repeated multiple times to build a multi-layer attention block. In our implementation, the number of heads for multi-head attention and the number of SCM attention layers are set to 4 and 2, respectively. We empirically found that the four layers and two heads can achieve the best performance in our experiment.

%\begin{figure}[!t]
%	\centering\
%	\includegraphics[width=\columnwidth]{./figures/figure4.pdf}
%	\caption{Architectural details of the proposed spatial and channel-wise multi-head attention (SCM-attention) block. $\oplus$ denotes elementwise summation.}
%	\label{fig:4}
%	\vspace{-0.2cm}
%\end{figure}

\subsection{Training and Instance Segmentation}
To train the end-to-end method for multi-class road defect detection and segmentation, we minimise a multi-task objective function consisting of the multi-class detection loss and  $L_{\text{Det}}$ and binary segmentation loss $\mathcal{L}^{\text{Seg}}$.

First, the multi-class detection loss $L_{\text{Det}}$ is defined as:
\begin{equation}
\label{eq:det_loss}
\begin{split}
L_{\text{Det}}(\{p_i\}, \{t_i\}) & = \frac{1}{N_{\text{anch}}}\sum_i\left(L_{\text{p-cls}}(p^*_i, p^{p}_i)+ L_{\text{m-cls}}(p_i,p^{c}_i)\right)\\ &+ \lambda\frac{1}{N_{\text{anch}}}\sum_i  p^{*}_i L_{\text{reg}}(t_i, t^{*}_i).
\end{split}
\end{equation}
Here, $i$ is the index of an anchor in each mini-batch, and $p_i$ is the predicted probability of anchor $i$ being a particular class's object. $L_{\text{p-cls}}$ is log loss related to object existence, so it is computed over two classes (object vs not object). The ground-truth labels related to object existence $p^{p}_i$ is one if the anchor is positive and 0 if the anchor is negative. $L_{\text{m-cls}}$ is applied for classifying defect classes so that the ground-truth labels for object classes $p^{c}_i$ are defined by a one-hot vector representing a particular object class. $p^*_i$ is defined by the maximum value among the predicted probability of the classification output. $t_i$ is a vector representing the four parameterised coordinates of the predicted bounding box, and $t^{*}_i$ is that of the ground-truth box. 

For the bounding box regression, we use $L_{\text{reg}}(t_i, t^{*}_i)=R(t_i - t^{*}_i)$ where $R$ is the robust loss function (smooth L$_1$). The term $p^{*}_i L_{\text{reg}}$ means the regression loss is activated only for positive anchors ($p^{*}_i=1$) and is disabled otherwise ($p^{*}_i=0$). Those loss terms are regularised by $N_{\text{cls}}$ and $N_{\text{reg}}$ and weighted by a balancing parameter $\lambda$. 

Second, the loss function for the segmentation head is defined by the combination of binary-cross entropy loss and dice loss, as follows:
\begin{equation}
\mathcal{L}_{\text{Seg}}(\{o_{i}\},\{\hat{o}_{i}\}) = \sum_i\left(L_{\text{bce}}(o_{i}, \hat{o}_{i})+ L_{\text{dice}}(o_{i},\hat{o}_{i})\right),
\label{eq:seg}
\end{equation}
where $o_{i}$ and $\hat{o}_{i}$ indicate the binary segmentation results and pixel-level annotation for each road defect. The reason why the defect division is divided into dual classes rather than multi-classes is as follows. As mentioned before, defects on the road surface have similar colours regardless of class, and if you try to distinguish them forcibly, the performance may deteriorate. In some cases, road defects are connected, and clearly distinguishing them is very difficult, even for experts with pixel-level labels.

In optimising the entire framework, except for the standard part that extracts the features corresponding to each head as input (the encoding part, See Fig. \ref{subfig:key-a}), the remaining parts (the network corresponding to each head) do not share the gradient of each loss function. Only the encoding part is jointly trained, and the detection and segmentation head are optimised independently. 

The instance segmentation process for multi-class road defects using SCM-MRCNN is performed by combining the detection and segmentation results of the segment head. The segmentation areas for multi-class road defects are divided based on the detection bounding box first. Then, the class assignment for each segment area has the highest probability among the detection results of the detection head ($p^{*}_{i}$ in Eq. \ref{eq:det_loss}), which is assigned to the output class. If several detection results exist, classes are allocated sequentially from the detection result with the highest class probability.

\section{Experimental results}
\label{sec:4}
\subsection{Dataset and evaluation metrics}
There is no benchmark for evaluating the multi-class road defect detection and segmentation performances; Therefore, we utilise defect segmentation datasets \cite{ozgenel2018performance} and road defect detection dataset \cite{arya2021rdd2020}, which are publicly available. Also, we created the RoadEYE dataset as a novel benchmark for multi-class road defect detection and segmentation. The descriptions of those datasets are as follows:

\noindent
\textbf{Crack Segmentation Dataset} is merged from 11 crack segmentation datasets \cite{liu2022datasets}. The dataset contains 11,200 images. All the images are resized to the size of 448 $\times$ 448, and include RGB channels. The two folders of images and masks contain all the images. The two folders, train and test, contain training and testing images split from the two above folders. The splitting is stratified so that the proportion of each dataset in the train and test folder is similar.

\noindent
\textbf{RDD2020 dataset} is created by \cite{arya2021rdd2020}, and it is a large-scale dataset that comprises 26,336 road images captured in India, Japan, and the Czech Republic. The dataset includes over 31,000 instances of road damage and is designed for developing deep learning-based methods to detect and classify various types of road damage automatically. RDD2020 covers a range of common road damage types, such as longitudinal cracks, transverse cracks, alligator cracks, and potholes. The dataset includes images with resolutions of either 600 $\times$ 600 or 720 $\times$ 720, providing a detailed view of the road surface.

\begin{figure}[!t]
	\centering
	\includegraphics[width=\columnwidth]{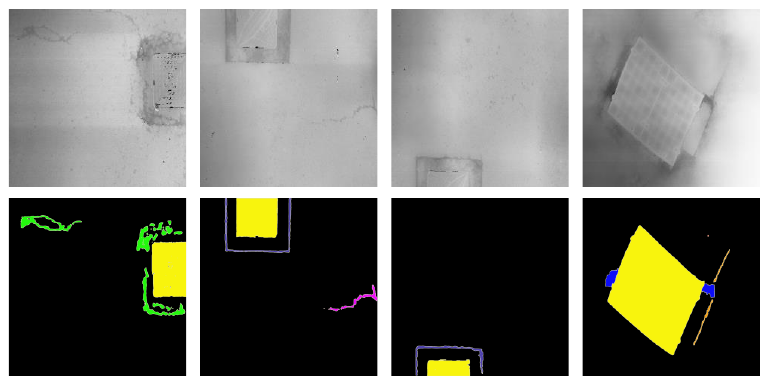}
	\caption{Example snapshots of a sample of the RoadEYE dataset. Top: images. Bottom: Annotation of instance segmentation for multi-class road defect detection and segmentation.}
	\label{fig:exp}
	\vspace{-0.2cm}
\end{figure}

\noindent
\textbf{RoadEYE dataset}\footnote{To access the dataset, please visit \url{https://github.com/Robotiz3d/dataset}} is a new benchmark for multi-class road defect detection and segmentation. We captured 1,404 road surface point clouds to create the dataset and generated grey-scaled images. We selected 407 images that contain road defects for labelling. Each sample of a road surface image, segmentation mask, and annotation file for detection bounding boxes. We classified numerous road defects into nine defect classes. Fig. \ref{fig:exp} and Table \ref{tbl:dataset} show the example snapshot of a sample and statistics of the RoadEYE dataset. The dataset is the first to provide segmentation masks and bounding box annotations for the six road defect classes.

This work performs performance evaluation in three areas: Binary-class segmentation, multi-class defect detection, and instance segmentation (\ie multi-class defect detection and segmentation). All defect classes are considered one class for evaluating the binary-class segmentation. Average Intersection over Union (AIU), Optimal Dataset Scale (ODS), and Optimal Image Scale (OIS) are used for the evaluation metrics. The multi-class defect detection's performance is evaluated using precision, recall, F1-score, and average precision (AP). The performance evaluation for the instant segmentation performance is conducted based on the average precision computed based on the segmentation mask (AP$_{\text{M}}$) and bounding boxes (AP$_{\text{B}}$).

\begin{table}[t]
\resizebox{\columnwidth}{!}{% 
\begin{tabular}{l|c|c|c|c|c|c|c}
\toprule
& Pothole & Manhole & Longitudinal  & Transverse  & Joint  & Wheel  & Total    \\
\midrule
\# & 253 & 172  & 182 & 101 & 184 & 248 & 1140  \\ %  (Eq. \ref{eq:general_daseg})
\hline
\% & 22.2 & 15.1  & 16.0 & 8.9 & 16.1 & 21.8  & 100  \\
\bottomrule
\end{tabular}
}
\caption{The proportions (\%) and number (\#) of instances corresponding to each defect class.} 
\vspace{-2ex}
\label{tbl:dataset}
\end{table}

\begin{table}[t]
\resizebox{\columnwidth}{!}{% 
\begin{tabular}{c|c|c|c|c|c|c|c|c}
\toprule
\midrule
 \multicolumn{3}{c|}{Naive-MRCNN} & \multicolumn{3}{c|}{SM-MRCNN} & \multicolumn{3}{c}{SCM-MRCNN}  \\
\hline
 AIU & AP$_{\text{M}}$  & AP$_{\text{B}}$  & AIU & AP$_{\text{M}}$  & AP$_{\text{B}}$   & AIU & AP$_{\text{M}}$  & AP$_{\text{B}}$     \\
\hline
\multicolumn{9}{c}{When $\mathcal{L}_{\text{Seg}}$ considers multi-class segmentation}\\
\hline
  4.7 & 32.5 & 62.5 & 7.4  & 51.6 & 69.3 & \bf{9.4}  & \bf{61.5} & \bf{72.7}\\ % (Eq. \ref{eq:general_daseg})
\hline
\multicolumn{9}{c}{When $\mathcal{L}_{\text{Seg}}$ considers binary-class segmentation}\\
\hline
  15.5 & 60.6 & 68.0 & 18.6  & 64.1 & 70.4 & \bf{22.6}  & \bf{68.6} & \bf{73.3} \\ %  (Eq. \ref{eq:general_daseg})
\midrule
\bottomrule
\end{tabular}
}
\caption{Performance evaluation results depending on architecture configuration and loss function settings on the RoadEYE dataset.} 
\vspace{-2ex}
\label{tbl:abl}
\end{table}

\begin{table*}[!htb]
    \begin{minipage}{0.83\columnwidth}
    \centering
    \resizebox{\textwidth}{!}{%
	\begin{tabular}{l c c c c c c}
		\toprule
		\midrule
		\multirow{2}[1]{*}{Methods} &
		\multicolumn{3}{c}{Crack Segmentation \cite{zou2012cracktree}} & 
		\multicolumn{3}{c}{RoadEYE} \\
		\cmidrule(l){2-4} \cmidrule(l){5-7} 
		&   AIU  & ODS  & OIS &  AIU & ODS  & OIS  \\
		\cmidrule(l){1-7}
          \multicolumn{1}{l}{MRCNN ($f_{\text{Seg}}$) \cite{liu2017richer}}  & 48.9 & 62.3  & 66.8  & 15.3 & 55.3 & 58.6  \\
		\multicolumn{1}{l}{RCF \cite{liu2017richer}}  & 46.7 & 53.2  & 60.3  & 19.6 & 51.7 & 60.2  \\
		\multicolumn{1}{l}{FCN \cite{long2015fully}}  & 38.1 & 50.9  & 58.6  & 10.8  & 51.6 & 41.9   \\
		\multicolumn{1}{l}{FPHBN \cite{yang2019feature}} & 50.1 & 61.7  & 65.0  & 10.3  & 53.3 & 55.4 \\
		\multicolumn{1}{l}{U-Net \cite{ronneberger2015u}} & 46.7 & 49.3  & 55.2  & 19.3  & 41.2 & 38.2 \\
  \multicolumn{1}{l}{DeepCrack \cite{yang2019feature}} & 45.6 & 59.1  & 53.2  & 17.6  & 49.2 & 27.6 \\
		\multicolumn{1}{l}{Dual-Swin \cite{liang2022cbnet}}  & 52.8 & 64.3  & 68.7  & 20.5 & 52.4 & 51.6 \\
		%\cmidrule(l){1-10}  
		\midrule
		\multicolumn{1}{l}{SCM-MRCNN ($f_{\text{Seg}}$)}  & \textbf{53.1} & \textbf{64.5}  & \textbf{69.0}  & \textbf{22.6} & \textbf{58.7} & \textbf{56.4} \\
		\midrule
		\bottomrule
	\end{tabular}
    }
	\caption{Quantitative comparison for the road defect segmentation performance using Crack Segmentation dataset and RoadEYE dataset. The \textbf{bolded} numbers indicate the best performance.}
 \label{table:2}
    \end{minipage}%
    \hspace{0.5ex}
    \begin{minipage}{1.20\columnwidth}
    \vspace{-3ex}
    \resizebox{\textwidth}{!}{%
   \begin{tabular}{l c c c c c c c c}
		\toprule
		\midrule
		\multirow{2}[1]{*}{Methods} &
		\multicolumn{4}{c}{RDD2020 \cite{arya2021rdd2020}} & 
		\multicolumn{4}{c}{RoadEYE} \\
		\cmidrule(l){2-5} \cmidrule(l){6-9} 
		&   Precision  & Reall  & F1 &  mAP & Precision  & Reall  & F1 &  mAP \\
		\cmidrule(l){1-9}
		\multicolumn{1}{l}{MRCNN ($f_{\text{Det}}$) \cite{liu2017richer}}  & 61.3 & 58.7  & 59.3  & 59.3 & 62.7 & 58.5  & 59.3& 61.7 \\
		\multicolumn{1}{l}{Faster-RCNN \cite{long2015fully}}  & 62.7 & 59.3  & 60.1  & 60.2  & 64.2 & 59.1 & 60.9 & 62.4  \\
		\multicolumn{1}{l}{Yolov5 \cite{yang2019feature}} & 58.5 & 57.3  & 57.8  & 57.2  & 58.5 & 55.3 & 55.9 & 57.8 \\
		\multicolumn{1}{l}{SSD \cite{ronneberger2015u}} & 57.9 & 57.1  & 57.3  & 56.3  & 60.9 & 56.0 & 56.4 & 57.5 \\
		\multicolumn{1}{l}{Zhang \etal~\cite{zhang2020exploring}}  & 58.3 & 57.5  & 56.4  & 56.5  & 61.7 & 57.7 & 58.9 & 60.2\\
		%\cmidrule(l){1-10}  
		\multicolumn{1}{l}{YOLO-LRDD~\cite{wan2022yolo}}  & 59.2 & 58.2  & 58.7  & 57.6 & 60.3 & 56.9 & 57.5 & 58.3 \\
		\midrule
		\multicolumn{1}{l}{SCM-MRCNN ($f_{\text{Det}}$)}  & \textbf{63.5} & \textbf{60.9}  & \textbf{61.5}  & \textbf{61.7} & \textbf{65.4} & \textbf{61.3}  & \textbf{62.7}  & \textbf{63.1}\\
		\midrule
		\bottomrule
	\end{tabular}
    }
	\caption{Quantitative comparison for the multi-class road defect detection performance using RDD2020 dataset \cite{arya2020global} and RoadEYE dataset. The \textbf{bolded} numbers indicate the best performance.}
    \label{table:3}
    \end{minipage}
    \begin{minipage}{\textwidth}
     \centering
      \vspace{2ex}
    \def\mrcnn{MRCNN \cite{he2017mask}}
    \def\fcis{Dual-Swin \cite{liang2022cbnet}}
    \def\msrcnn{Centermask~\cite{lee2020centermask}}
    \def\retinamask{Mask Transfiner~\cite{ke2022mask}}
    \def\masklab{YOLACT~\cite{bolya2019yolact}} 
    \newcommand{\modelname}[1]{\methodname{}-#1}
     \resizebox{\textwidth}{!}{%
    \begin{smalltable}{l c c c c c c c c c c c c c c c c c  } \toprule
        \multirow{2}[1]{*}{Method}         &    \multirow{2}[1]{*}{Parms}    &   \multirow{2}[1]{*}{FPS}   &    \multicolumn{2}{c}{Pothole}     &  \multicolumn{2}{c}{Manhole} & \multicolumn{2}{c}{Longitudinal} &  \multicolumn{2}{c}{Transverse} & \multicolumn{2}{c}{Joint}  &  \multicolumn{2}{c}{Wheel}  &  \multicolumn{2}{c}{mAP} \\
        \cmidrule(l){4-5} \cmidrule(l){6-7} \cmidrule(l){8-9} \cmidrule(l){10-11} \cmidrule(l){12-13} \cmidrule(l){14-15} \cmidrule(l){16-17} 
		&  & &  AP$_{\text{M}}$ &  AP$_{\text{B}}$  & AP$_{\text{M}}$ &  AP$_{\text{B}}$ & AP$_{\text{M}}$ &  AP$_{\text{B}}$ & AP$_{\text{M}}$ &  AP$_{\text{B}}$ & AP$_{\text{M}}$ &  AP$_{\text{B}}$  & AP$_{\text{M}}$ &  AP$_{\text{B}}$& AP$_{\text{M}}$ &  AP$_{\text{B}}$  \\
        \midrule
        \mrcnn          &    \bf{44 M}  &    11.7 &  77.8 & 86.3  &      74.2 &  82.1  & 65.2 &  76.4   &   54.2 &  60.5 & 38.2 & 42.3 & 54.0 &  60.3  & 60.6 & 68.0\\
        \retinamask     &    \bf{44 M}    &   13.1 & 81.2 &  84.6 &      \bf{79.4} &  81.6  & 58.1&   62.3  & 52.3 &  58.2  & 40.2 & 43.6 & 55.2 & 58.2  & 61.1 & 64.8\\
        \fcis           &   107 M    &    7.4 & 84.1 & 90.4  & 76.6 &   84.3   &   75.3 & 85.2 &   \bf{57.2}  &  \bf{61.8} &    38.1 & 39.6 & 58.8 & 59.2 & 65.1 & 70.1 \\
        \masklab           &    114 M   &     \bf{21.7} & 67.3 &  69.3     &      71.1 & 73.2    & 61.5 &  66.7   &      42.7  & 46.3     & 21.7 & 22.8 & 36.2 & 40.1 & 50.1 & 53.1\\
        \msrcnn         &    114 M   &     8.4 & 73.0 & 79.5 &  75.2 &  80.2    & 76.1 &  80.9     &     51.4  & 56.9 & 20.6 & 23.5 & 47.1 & 48.2 & 57.2 & 61.5\\
        \midrule
        He \etal~\cite{he2022pavement}  &   65 M   & 8.5 & 72.4 & 74.2 &     73.0 &  76.3   &  62.1 &  65.4  &    44.8 & 48.1 & 27.5 & 30.6&51.7 & 55.7& 55.2 & 58.3\\
        Zhang \etal~\cite{zhang2022multi}         &   65 M     &   8.1  &  71.8 & 73.2 &   74.5 &   76.9   &    64.5 &   68.4    &      48.2 & 50.8 & 27.5 &  31.6 & 50.6 & 55.7 & 56.2 & 59.4\\
        PGA-NET~\cite{dong2019pga}         &    45 M    &  9.8 & 82.6 & 88.2  &     77.0  &   82.5    &     74.8  &  80.2   &      54.1 & 59.4 & 46.0 & 49.2 & 58.3 & 61.2 & 65.5 & 70.1\\
        \midrule
       SCM-MRCNN &  53 M  & 9.1 &  \bf{84.2} &   \bf{91.3}  &  79.3    &     \bf{85.4} &   \bf{81.2}    & \bf{86.8} &   56.9     &      61.5 & \bf{50.3}   &  \bf{54.0}  &  \bf{59.7}   &  \bf{63.0} & \bf{68.6}   &  \bf{73.3}\\
        \bottomrule
    \end{smalltable}
    }
    \vspace{-0.05in}
    \label{tbl:4}
    \caption{Quantitative comparison for the multi-class road defect detection and segmentation performance using RoadEYE dataset. The \textbf{bolded} numbers indicate the best performance.}
    \label{table:4}
    \end{minipage}
\end{table*}

\subsection{Implementation}
The segmentation loss $\mathcal{L}_{\text{Seg}}$ is defined only for positive detection results. The mask target is the intersection between a RoI and its associated ground truth. The results will be considered positive when IoU between a detected bounding box and corresponding annotation is over 0.5. The resolution of images is resized to 224 $\times$ 224. A stochastic gradient descent (SGD) is used for optimising parameters. The batch size is set by 4. To obtain experimental results for the three datasets, we train each dataset for 200 epochs, with a starting learning rate of 0.01. Learning rate decay is applied so that it is decreased by multiplying it by 0.99 for every 20 epoch.

\subsection{Ablation study}
The SCM-MRCNN has two remarkable changes compared with the original MRCNN \cite{he2017mask}. First, in training the SCM-MRCNN, we applied a binary segmentation strategy for the segmentation head ($f_{\text{Seg}}$). Instead of cross-entropy loss for multi-class segmentation, we applied combination loss of the binary cross entropy and dice loss (See Eq. \ref{eq:seg}). Second, as shown in Fig. \ref{subfig:key-a}, the SCM attention blocks have been embedded between each convolutional or deconvolutional layer to learn a robust representation spatially and channel-wise. We have conducted ablation studies to demonstrate the effectiveness of those critical changes.

Table \ref{tbl:abl} presents the evaluation results depending on loss function settings and architectural configurations. Training a segmentation head using a binary-class segmentation setting is more beneficial for multi-class road defect detection and segmentation. The MRCNNs trained by the binary-class segmentation outperform counterparts with significant performance gaps \textit{e.g.,} Naive-MRCNN trained by the multi-class segmentation settings produces 4.7 AIU and 32.5 AP$_{\text{M}}$; however, it obtains 15.5 AIU and 60.6 AP$_{\text{M}}$ when the binary-class segmentation setting trains it.

In terms of the performance analysis concerning the architectural configuration, we compare the average precision computed based on segmentation mask (AP$_{\text{M}}$) and detection bounding box (AP$_{\text{B}}$). MRCNN without structural modification (Naive-MRCNN) achieves AP$_{\text{M}}$ of 60.6 and AP$_{\text{B}}$ of 68.0. These are the lowest performance. MRCNN with spatial multi-head attention (SM-MRCNN) and the MRCNN with SCM attention block (SCM-MRCNN) achieve better performance. 

% \begin{figure}[!t]
% 	\centering
% 	\includegraphics[width=\columnwidth]{./figures/fig6.pdf}
% 	\caption{The qualitative comparison on the KCL-R3 dataset.}
% 	\label{fig:results}
% 	\vspace{-0.2cm}
% \end{figure}

\subsection{Comparison with existing methods}
The proposed SCM-MRCNN can provide three outputs: 1) Multi-class detection, 2) Binary segmentation, and 3) instance segmentation. We compare the performance of our method in terms of all the aforementioned aspects, respectively. The methods we have chosen for comparison include detection and segmentation methods for road defects and building defects \cite{liu2017richer,long2015fully,yang2019feature,wan2022yolo} in a similar domain. Additionally, because our method can be classified as an instant segmentation method, we have compared our method with various SOTA instance segmentation methods \cite{liang2022cbnet,lee2020centermask,ke2022mask,bolya2019yolact}.

Table \ref{table:2} compares defect segmentation performance using the CS and RoadEYE datasets. The SCM-MRCNN produces AID of 53.1, ODS of 64.5, and 69.0 OIS on the CS dataset. For the RoadEYE dataset, SCM-MRCNN produces AID of 22.6, ODS of 58.7, and 56.4 ODS on the CS dataset. These figures are the best performances in our experiments. Dual-Swin \cite{liang2022cbnet} produces the second-ranked performance. Interestingly, Dual-Swin also uses a self-attention module to learn a long-term dependency.  Both Dual-Swin and SCM-MRCNN can learn features with spatially long-term dependencies through self-attention. These experimental results suggest that long-term dependencies help segment road defects.

Table \ref{table:3}  compares multi-class defect detection performance using the RDD2020 and RoadEYE datasets. The SCM-MRCNN produces 63.5 precision, 60.9 recall, 61.5 F1-score, and 61.7 mAP on RDD2020 dataset. For the RoadEYE dataset, SCM-MRCNN produces 65.4 precision, 61.3 recall, 62.7 F1-score, and 63.1 mAP. These figures are the first-ranked performances in our experiments. Faster-RCNN \cite{long2015fully} produces the second-ranked performance. Noticably, Region proposal-based methods \cite{he2017mask,long2015fully} including the proposed SCM-MRCNN show better performance than single-shot object detection-based methods \cite{wan2022yolo,ronneberger2015u}. This suggests that the region-proposal modules can cover overall geometric information so that it helps identify various types of defects.

Performance comparisons with recent SOTA multi-class road defection and segmentation and instant segmentation methods are presented in Table \ref{table:4}. The SCM-MRCNN produces AP$_{\text{M}}$ of 68.6, AP$_{\text{B}}$ of 73.3. These figures are the best performances in our experiments. However, for the Transverse crack class, the SCM-MRCNN produces second-ranked performances. Dual-Swin \cite{liang2022cbnet} produces better performance. The performance comparison results show that the SCM-MRCNN outperforms SOTA methods for road defect detection and instance segmentation. 

For computational cost, the SCM-MRCNN has 53 million (M) parameters. This model size is relatively bigger than other MRCNN-based methods \cite{he2017mask,ke2022mask} since the SCM-attention blocks have been embedded between each convolutional or deconvolutional layer, which make the network structure more complicated. However, it is smaller than YOLACT \cite{bolya2019yolact} and Centermask \cite{lee2020centermask}, which have over 100 M parameters. The method shows the fastest execution speed is YOLACT. The YOLACT shows 21.7 FPS. The SCM-MRCNN achieves 9.1 FPS, which is relatively slower than the YOLACT. However, the SCM-MRCNN obtains faster execution speed compared with Dual-Swin \cite{liang2022cbnet}, and Centermask \cite{ke2022mask}.
 
% \subsection{Computational complexity and execution speed}
% Our method's computational complexity will vary depending on the baseline method's complexity and the number of source domains. The computational complexity of our method, which is estimated based on the model complexity study of convolutional neural networks \cite{he2015convolutional}, is $O(n^{\text{s}}\sum_{i=1}^{d}W_{i}H_{i}D_{i}w_{i}h_{i}d_{i})$, where $n^{\text{s}}$ denotes the number of source domains. $d$ is the number of convolutional layers. $W_{i}$, $H_{i}$, and $D_{i}$ indicate the width, height, and depth of input data and the outputs of each convolutional layer. $w_{i}$, $h_{i}$, and $d_{i}$ define the width, height, and the using kernel of each convolutional layer. In our experiments, our method takes 0.167 seconds to process one image. It is 0.042 seconds longer than that of the U-Net \cite{ronneberger2015u}, which is the baseline model of our method. This figure is obtained by applying parallel processing, and the processing speed of the proposed method is proportional to the number of source domains used. 

\section{Conclusion}
\label{sec:5}
This paper presents an end-to-end framework for multi-class road defect detection and segmentation. To detect and segment multi-class road defects based on a single framework, it is essential to derive spatially and channel-wisely generalised representation. We design the spatial and channel-wise multi-head attention (SCM Attention) to learn better-generalised representation across all dimensions. The experimental results demonstrated the effectiveness of the SCM attention and the SCM-MRCNN for multi-class road defect detection and segmentation.

\small
\bibliographystyle{IEEEtran}
\bibliography{icra_24_bib}

\end{document}